%% file: main.tex
\documentclass[10pt,journal,compsoc]{IEEEtran}

\usepackage{caption}
\usepackage{times}
\usepackage{epsfig}
\usepackage{graphicx}
\usepackage{amsmath}
\usepackage{amssymb}
\usepackage{multirow}
\usepackage{footnote}
\usepackage[capitalize]{cleveref}
\usepackage{url}
\usepackage{xcolor,colortbl}
\usepackage{enumitem}
\ifCLASSOPTIONcompsoc
  \usepackage[nocompress]{cite}
\else
  \usepackage{cite}
\fi
\ifCLASSINFOpdf
\else
\fi
\begin{document}
\title{Augmentation Pathways Network \\for Visual Recognition}

\author{Yalong~Bai,%~\IEEEmembership{Member,~IEEE,}
        ~Mohan~Zhou,%~\IEEEmembership{Fellow,~OSA,}
        ~Wei Zhang,
        Bowen Zhou,~\IEEEmembership{~Fellow,~IEEE}
        and~Tao~Mei,~\IEEEmembership{~Fellow,~IEEE}% <-this % stops a space
\IEEEcompsocitemizethanks{\IEEEcompsocthanksitem Yalong Bai, Wei Zhang, Bowen Zhou, Tao Mei are with JD Explore Academy, Beijing, China, 100010.\protect\\
E-mail: ylbai@outlook.com
\IEEEcompsocthanksitem Mohan Zhou is with Harbin Institute of Technology.}% <-this % stops an unwanted space
}


\markboth{Journal of \LaTeX\ Class Files,~Vol.~14, No.~8, August~2015}%
{Shell \MakeLowercase{\textit{et al.}}: Bare Demo of IEEEtran.cls for Computer Society Journals}
\IEEEtitleabstractindextext{%
\input pami_src/abstract
}

\maketitle

\IEEEdisplaynontitleabstractindextext

% \fi
\IEEEpeerreviewmaketitle

\input pami_src/introduction

\input pami_src/related_work
\input pami_src/method
\input pami_src/experiments
\input pami_src/conclusion
\section*{Acknowledgments}
This work was supported by the National Key R\&D Program of China under Grand No.2020AAA0103800.

\ifCLASSOPTIONcaptionsoff
  \newpage
\fi



\bibliographystyle{IEEEtran}
\bibliography{egbib.bib}
\begin{IEEEbiographynophoto}{Yalong Bai}
is a Senior Researcher at JD.com. He received his Ph.D. degree in Harbin Institute of Technology and Microsoft Research Asia Joint Ph.D. Education Program at 2018. His research interests include representation learning, multimodal retrieval, visual question answering, and visual commonsense reasoning. He has won first place in several international challenges on CVPR, ICME and MM. He has also served as the Area Chair for ACM MM Challenge and ICASSP.
\end{IEEEbiographynophoto}
\begin{IEEEbiographynophoto}{Mohan Zhou}
is currently a Ph.D. student at Harbin Institute of Technology, China, under the supervision of Prof. Tiejun Zhao. Meanwhile, he also works as a research intern at JD Explore Academy. Before that, he received his B.Eng. degree also from Harbin Institute of Technology in 2021. His current research interest includes representation learning and multilearning. He achieved impressive results in several fine-grained image classification competitions organized by CVPR workshop.
\end{IEEEbiographynophoto}
\begin{IEEEbiographynophoto}{Wei Zhang}
is now a Senior Researcher at JD.com, Beijing, China. He received his Ph.D degree from the Department of Computer Science in City University of Hong Kong. His research interests include computer vision and multimedia, especially visual recognition and generation. He has won the Best Demo Awards in ACM MM 2021, and served as the Area Chair for ICME, ICASSP, and Technical Program Chair for ACM MM Asia 2023.
\end{IEEEbiographynophoto}
\begin{IEEEbiographynophoto}{Bowen Zhou}
(Fellow, IEEE) has been the President of Artificial Intelligence Platform \& Research of JD.com since September 2017. Bowen is a technologist and business leader of human language technologies, machine learning, and artificial intelligence. Prior to joining JD.com, Dr. Zhou held several key leadership positions during his 15-year tenure at IBM Research’s headquarters. He previously served as a member of the IEEE Speech and Language Technical Committee, Associate Editor of IEEE Transactions, ICASSP Area Chair (2011-2015), ACL, and NAACL Area Chair.
\end{IEEEbiographynophoto}
\begin{IEEEbiographynophoto}{Tao Mei}
(Fellow, IEEE) is a vice president with JD.COM and the deputy managing director of JD Explore Academy, where he also serves as the director of Computer Vision and Multimedia Lab. Prior to joining JD.COM in 2018, he was a senior research manager with Microsoft Research Asia in Beijing, China. He has authored or coauthored more than 200 publications (with 12 best paper awards) in journals and conferences, 10 book chapters, and edited five books. He holds more than 25 U.S. and international patents. He is a fellow of IAPR (2016), a distinguished scientist of ACM (2016), and a distinguished Industry Speaker of IEEE Signal Processing Society (2017).
\end{IEEEbiographynophoto}
\end{document}

%% file: pami_src/abstract.tex
\begin{abstract}
Data augmentation is practically helpful for visual recognition, especially at the time of data scarcity. However, such success is only limited to quite a few light augmentations (e.g., random crop, flip). Heavy augmentations are either unstable or show adverse effects during training, owing to the big gap between the original and augmented images. This paper introduces a novel network design, noted as Augmentation Pathways (AP), to systematically stabilize training on a much wider range of augmentation policies. Notably, AP tames various heavy data augmentations and stably boosts performance without a careful selection among augmentation policies. Unlike traditional single pathway, augmented images are processed in different neural paths. The main pathway handles the light augmentations, while other pathways focus on the heavier augmentations. By interacting with multiple paths in a dependent manner, the backbone network robustly learns from shared visual patterns among augmentations, and suppresses the side effect of heavy augmentations at the same time. Furthermore, we extend AP to high-order versions for high-order scenarios, demonstrating its robustness and flexibility in practical usage. Experimental results on ImageNet demonstrate the compatibility and effectiveness on a much wider range of augmentations, while consuming fewer parameters and lower computational costs at inference time.
\end{abstract}

% need definition of light / heavy

%% file: pami_src/introduction.tex
\IEEEraisesectionheading{\section{Introduction}\label{sec:introduction}}

% 需要新的network design来区别对待heavy aug。收割有用的信息，抑制有害信息。

% 做法：
% 1. replace one path way with two pathway。
% 2. replace one conv with two conv；分别处理不同数据来源的chanel；（replace one channel with two separate channels）

% ours：aug-specific representation 1.用invariant data去处理 ... 

% we argue: 1. light aug should go for traditional aug-invariant paradigm, 2. while heavy aug are better going to our nested conv(?). 
% harness the data aug. go with separate pathway, with different 

% Unfortunately, many application domains do not have access to big data, such as medical image analysis. 

% these networks are heavily reliant on big data to avoid overfitting.

% improve the performance of their models and expand limited datasets to take advantage of the capabilities of big data.

% https://arxiv.org/pdf/1906.04547.pdf
% data augmentation invariance, an unsupervised learning objective which improves the robustness of the learned representations by promoting the similarity between the activations of augmented image samples.
% 这里提供了data augmentation invariance

% augmentation types:
% light: identity (original image), flip, crop, ...
% heavey: grayscale, gridshuffle, 

% aug介绍：reduce CNN generalization error by simulating realistic variations of training data，but sometimes，aug引入了新的gap

% artificial varitions are hand-enginnered to mimic the appearance feature of future test samples that derive from the training manifold. 
\IEEEPARstart{D}{eep} convolutional neural networks (CNN) have achieved remarkable progress on visual recognition. In some cases, deep models are likely to overfit the training data as well as its noisy signals~\cite{zhang2016understanding}, even on a large-scale dataset such as ImageNet~\cite{krizhevsky2012imagenet,srivastava2014dropout}. Data augmentation usually serves as a standard technique for regularizing the training process and reducing the generalization error, especially when data annotations are scarce. 
% By learning a augmentation-invariant feature representation, convolutional neural networks can benefits from augmented data for learning a more robust feature.
%By generating ``safe" sample variants, data augmentation is considered as giving more samples in the feature space. 
% Thus data augmentation is usually treated as a direct way boosting the model performance and robustness for free, especially when data annotations are scarce.

% motivation
% light aug经常有用，但heavy aug比较unstable
% 原因：gap among 《train/aug/test data》
% 1. aug data与test data gap太大，导致效果不好==>分类器学习偏移
% 2. 另一个原因：aug data与train set的gap大，（默认train、test的gap小）==>导致分类器学习偏移

\begin{figure*}[!t]
  \centering
  \includegraphics[width=0.95\linewidth,page=1]{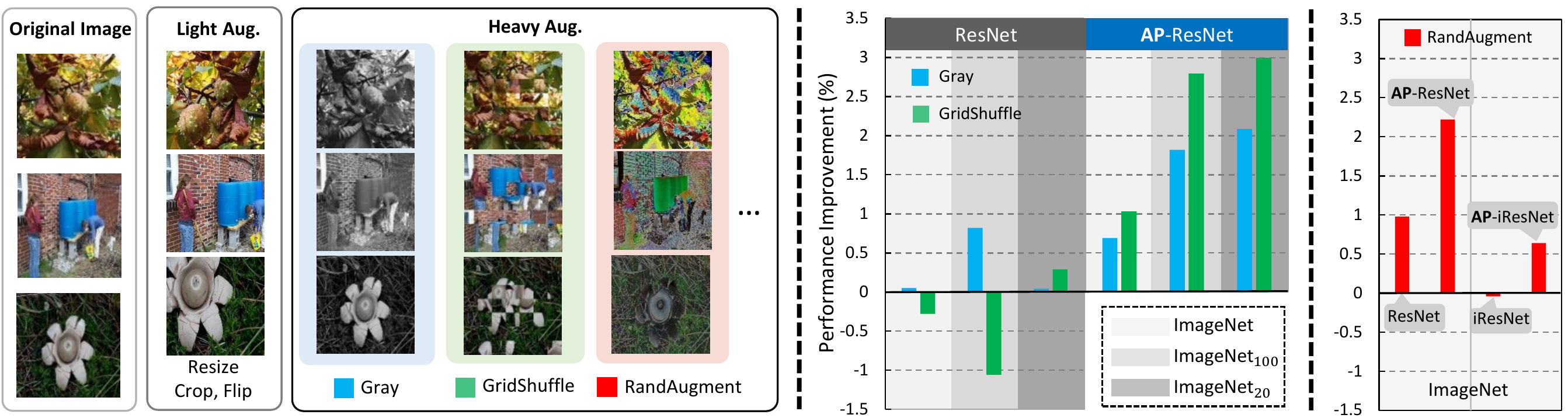}
  \caption{\textbf{Left}: Examples of original images and their lightly augmented (randomly Resize, Crop, Flip) and heavily augmented (Gray, GridShuffle, RandAugment) versions. \textbf{Middle}: Improvement on Top-1 accuracy by applying two heavy augmentations (Gray and GridShuffle) on ImageNet and its subsets (ImageNet$_n$, $n$ indicates the number of images used per category). Standard network (ResNet-50) performs quite unstable, showing marginal or adverse effects. \textbf{Right}: Improvement on Top-1 accuracy by applying searched augmentation (RandAugment~\cite{cubuk2020randaugment}: A collection of randomly selected heavy augmentations) on ImageNet. Augmentation policy searched for ResNet-50 leads to performance drop on iResNet-50. In contrast, Augmentation Pathways (AP) based network can steadily benefit from a much wider range of augmentation policies for robust classification.}
  \label{fig:motivation}
\end{figure*}

However, such successes in data augmentation are only restricted to a handful of augmentations that slightly jitters the original image. 
%Other augmentation policies are proposed for some specific datasets, 
A large collection of augmentation operations can not be easily applied to arbitrary configurations (e.g., datasets, backbones, hyper-parameters). Sometimes data augmentation only shows marginal or even adverse effects on image classification. 
%Such unstable performances based on different datasets and hyperparameters. %, e.g. $\texttt{GridShuffle}$ (as shown in Figure~\ref{fig:intro}), which aims to destruct the object structures in images, can significantly improve the performance on fine-grained recognition tasks~\cite{chen2019destruction}, but results in performance drop in ILSVRC-2012 dataset~\cite{deng2009imagenet} as we shown in Table~\ref{tab:motivation}. 
Following the definition in prior works (e.g., SimCLR~\cite{chen2020simple}, imgaug toolkit~\cite{imgaug}, DSSL~\cite{bai2022directional}), we roughly group 
%To study this problem, we follow the convention by several prior works, including the popular ``imgaug''~\cite{imgaug} toolkit and SimCLR~\cite{chen2020simple}, where ``heavy augmentation usually leads to significant changes or information loss in images", and roughly group 
augmentation operations into two categories (Fig.~\ref{fig:motivation} left). 1) \textit{Light Augmentation} that only slightly modifies an image without significant information loss. Typical operations include random Flip, Crop~\cite{krizhevsky2012imagenet,simonyan2014very,Xie2016,li2019selective}. Note that the original image can also be treated as a special case of light augmentation (i.e., Identity). 2) \textit{Heavy Augmentation} (or named Strong Augmentation~\cite{wang2022contrastive}) that largely alters the image appearance, sometimes striping out a significant amount of information (such as color, object structure). Typical operations include Gray (transforming color image to grayscale), GridShuffle~\cite{chen2019destruction} (destructing object structures by shuffling image grids) and CutOut~\cite{devries2017improved} (masking out random area of image), etc.

%%% 1. 对于manually designed得DA: 不稳定，可能有hurt
%%% 2. 对于search得DA: 可能只适用于特定得network architecture。（randaug： resnet:iresnet）

Based on prior studies~\cite{krizhevsky2012imagenet,simonyan2014very,he2016deep}, light augmentations have demonstrated stable performance improvements, since lightly augmented images usually share very similar visual patterns with the original ones. However, heavy augmentations inevitably introduce noisy feature patterns, following different distributions with the original samples. Thus training directly with these images are often unstable, sometimes showing adverse effect in performance. For example in Fig.~\ref{fig:motivation} (Middle), GridShuffle is highly unstable on ImageNet, if trained with standard network (see ResNet column). 
This may be due to the implicit gap among three sets of ``train, augmented, test'' data. 
% 这句可以下移。
%The design principle of prior works is to lean an augmentation-invariant feature, i.e., accepting all information in augmented data as ``good". However, such assumption is not always true, especially on heavy augmentations. 
% 下面这句需要修改
% However, if we apply $\texttt{GridShuffle}$ on a subset of ImageNet (named ImageNet$_{20}$, randomly sampling 20 images per class), the top-1 accuracy increases. It means that there is still room for mining useful information from destructive augmentations.
%
%following the point above: data augmentation contains useful information. the key is how to use it.
%
% 目前针对aug data，网络architecture、输入方式、loss设计的principle都是学个aug-invariant feature，ie，sharing the same feature, with and without aug. 
% 好的坏的都一视同仁，接受所有的data,未区分数据中有用的，没用的（误导的）信息。

% a ref augmentation is important：
% https://ccneuro.org/2018/proceedings/1046.pdf 
Intuitively, heavy augmentations also introduce helpful and complementary information during training~\cite{chen2019destruction}. Recent studies~\cite{hernandez2018deep,hernandez2019learning} also suggest that networks trained with heavier augmentation yield representations that are more similar between deep neural networks and human brain. 
% The question still remains on how to design suitable network to fit.
However, heavy augmentation tends to generate images with larger variations from the original feature space. Such variations are not always helpful, since irrelevant feature bias is also introduced alongside the augmentation. From the opposite view, there is still useful information implied in the shared visual patterns between the original and heavily augmented images.
%Recall that light data augmentations steadily boost the performance. 
%The critical element to effective data augmentation is expected to be implied in the shared visual patterns between the original image and augmented image.
For example, contour information is augmented, but color bias is introduced in Gray augmentation; visual details are augmented, while object structure is destroyed in GridShuffle augmentation~\cite{chen2019destruction}. % heavy augmentation 有用，但做不好。
Therefore, expertise and knowledge are required to select feasible data augmentation policies~\cite{chen2019destruction}. In most cases, this is quite cumbersome. Even when augmentation improvements have been found for one specific domain or dataset, they often do not transfer well to other datasets. Some previous works employ search algorithms or adversarial learning to automatically find suitable augmentation policies~\cite{cubuk2019autoaugment, lim2019fast, hataya2019faster, cubuk2020randaugment}. However, such methods require additional computation to obtain suitable policies. Moreover, augmentation policies searched for one setting are usually difficult to fit other settings. For example in Fig.~\ref{fig:motivation} (Right), RandAugment~\cite{cubuk2020randaugment} searched for ResNet leads to slight performance drop in iResNet~\cite{duta2020improved} (an information flow version of ResNet).
%As an alternative to standard convolutional layer, AP-Conv stabilizes the performance improvement, and show  robustly from data augmentations by proposing another augmentation pathway. 

\begin{figure}[!t]
  \centering
  \includegraphics[width=0.9\linewidth,page=1]{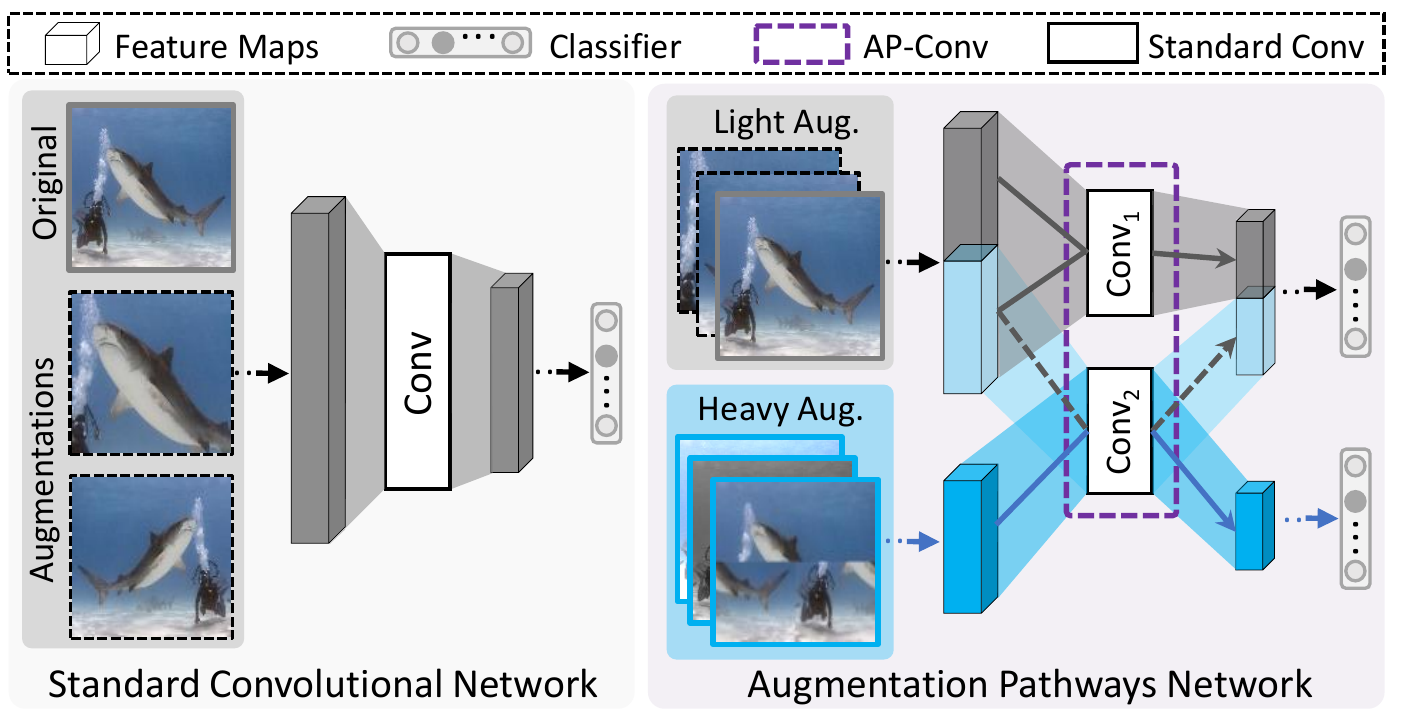}
  \caption{Illustration of standard CNN (Left) and our proposed Augmentation Pathways network (Right) for handling data augmentations. 
  %The 3rd-order extension (Right) is expanded from the basic AP-Conv but can handle heavy augmentations under two selections of hyperparameter according to the visual feature dependencies among input images. 
  Details of the basic AP-Conv in purple dashed box is illustrated in Fig.~\ref{fig:basicnestedconv}.}
\label{fig:intro}
\end{figure}

In this work, we design a network architecture to handle a wide range of data augmentation policies, rather than adapt augmentation policies for specific datasets or architectures. A plug-and-play ``Augmentation Pathways'' (AP) is proposed for restructuring the neural paths by discriminating different augmentation policies. Specifically, a novel augmentation pathway based convolution layer (AP-Conv) is designed to replace standard Conv layer to stabilize training with a wide range of augmentations. 
As an alternative to the standard convolutional layer, AP-Conv adapts network design to a much wider range of heavy data augmentations. As illustrated in Fig.~\ref{fig:intro}, traditional convolutional neural networks directly feed all images into the same model. In contrast, our AP-Conv (right of Fig.~\ref{fig:intro}) process the lightly and heavily augmented images through different neural pathways. Precisely, a basic AP-Conv layer consists of two convolutional pathways: 1) the main pathway focuses on light augmentations, and 2) the augmentation path is shared among lightly and heavily augmented images for learning common representations for recognition. Two pathways interact with each other through the shared feature channels. To further regularize the feature space, we also propose an orthogonal constraint to decouple features learned from different pathways. Notably, our AP-Conv highlights the beneficial information shared between pathways and suppresses negative variations from heavy data augmentation. In this way, the Augmentation Pathways network can be naturally adapted to different data augmentation policies, including manually designed and auto-searched augmentations.
%domain-specific features of lightly augmented images.  Finally, the common visual patterns and the in-domain visual patterns are united for optimizing objective function and result in a significant performance on original image recognition.

Furthermore, different augmentation hyperparameters may lead to different visual appearances and classification accuracy. Tuning such hyperparameters is non-trivial. Some works propose to automatically search for a proper hyperparameter. However, these methods usually require additional computation or searching cost~\cite{cubuk2019autoaugment}, and the learned augmentation policies are dataset or network dependent~\cite{lim2019fast,ho2019pba}. Thus these methods are usually with limited generalization capability. To address this, we gather all useful information from one augmentation policy with various hyperparameters, instead of selecting one most appropriate hyperparameter as previous works did. Specifically, we extend the augmentation pathways into high-order for processing training data from multiple hyperparameter selections of data augmentation pass different pathways. In this way, the information dependencies among different hyperparameters of data augmentation policies can be well structured, and the information from different neural network pathways can be gathered to organize a well-structured and rich feature space.

Comparing to the standard convolutional layer, our AP-Conv contains fewer connections and parameters. Moreover, it is highly compatible with standard networks. AP-Conv based network can even be directly finetuned from the standard CNN. The experimental results on ImageNet dataset demonstrated AP-Conv's efficiency and effectiveness by equipping manually designed heavy augmentations and the searched data augmentations collection.

%% file: pami_src/related_work.tex
\section{Related Work}
%Data augmentation plays an essential role in deep neural network based vision tasks, including image recognition, detection and segmentation. 
\noindent\textbf{Manually designed augmentation}\quad Since data augmentation can increase the training data diversity without collecting additional samples, it usually plays an essential role in deep neural network based vision tasks and benefits the model generalization capability and performance improvement as a standard operation in deep vision model training. In general, light data augmentation policies, including random cropping, horizontal flips are commonly used in various tasks~\cite{he2016deep,krizhevsky2017imagenet,chen2019mmdetection,girshick2018detectron}. Such data augmentation methods keep the augmented images in the original training set and lead to steady performance improvement in different neural network architectures trained on various datasets. Recently, heavy data augmentation methods have received more attention from the computer vision research community. Some methods~\cite{devries2017improved,zhang2017mixup,yun2019cutmix} randomly erase image patches from the original image or replace the patches with random noise. GridShuffle~\cite{chen2019destruction} is proposed for destructing the global structure of the object in images and force the model to learn local detail features. However, such manually designed heavy data augmentation is dataset-specific and usually suffer from adapting to different datasets. 

\noindent\textbf{Searched augmentation}\quad Inspired by the successes of Neural Architecture Search algorithms on various computer vision tasks~\cite{real2019regularized,zoph2018learning}, there are several current studies proposed for automatically search algorithms to obtain augmentation policies for given datasets and network architectures. These studies try to find the best augmentation policy collection from the predefined transformation functions by RL based strategy~\cite{cubuk2019autoaugment}, Population based training~\cite{ho2019pba}, Bayesian optimization~\cite{lim2019fast} or the latest grid search based algorithms~\cite{cubuk2020randaugment}. Such methods usually takes lots of GPU hours for searching a 
proper data augmentation collection before training model. Moreover, theoretically, these data augmentation strategies are dataset specific and network architecture specific. These two limitations hurt the practical value of the searched-based data augmentation methods.

%In the paper, we introduce a new viewpoint for the inter-dependency among dataset, network architecture, and data augmentation policies. Rather than selecting proper data augmentation policies for each dataset or network architecture, we propose a network architecture design for dealing with various data augmentations, including not only the light augmentation, but also manually designed heavy augmentation and the auto-searched data augmentation combinations. Our proposed augmentation pathways network can seek common ground from different kinds of data augmentation methods while reserving the differences among them. With lower training/inference computational cost, our method can achieve stable performance improvements on various network architectures and datasets equipping different kinds of data augmentation methods.
In the paper, we introduce a new viewpoint for the inter-dependency among dataset, network architecture, and data augmentation policies. Rather than selecting proper data augmentation policies for each dataset or network architecture, we propose a network architecture design method for dealing with various data augmentations, including not only the manually designed augmentation but also searched augmentation. With lower computational cost, our method can achieve stable performance improvements on various network architectures and datasets equipping different kinds of data augmentation methods.

%% file: pami_src/method.tex
\section{Methodology}
%This section presents the details of our augmentation pathway. 
In this section, we start with a general description of the basic augmentation pathway (AP) network (Sec.~\ref{ssec:basic}), then introduce two extensions of AP (Sec.~\ref{ssec:extension}) for handling multiple hyper-parameters of given augmentation policy.

We focus on deep convolutional neural network (CNN) based fully supervised image classification problem. A typical CNN architecture consists of $T$ stacked convolutional layers $\{c_1, c_2, ..., c_T\}$, and a classifier $f$. Given training image $I_i$ with its category label $l_i$, $\phi_i$ denotes the lightly augmented version of $I_i$. Note that the original input image $I$ can be regarded as a special case of $\phi$. The overall objective of a typical image classification network is to minimize:
\begin{equation}
\begin{split}
    \mathcal{L}_{cls} &= \sum_{i=1}^{N}\mathcal{L}\left(f(c_{T}(\phi_i)), l_i\right),
    %c_{t}(\phi_i)&=W_t c_{t-1}(\phi_i) + b_t,
\end{split}
\end{equation}
where $c_{t}(\phi_i)=W_t c_{t-1}(\phi_i) + b_t$, $\mathcal{L}$ is the cross-entropy loss, $W_t\in \mathbb{R}^{n_{t-1}\!\times\!h_{t}\!\times\!w_{t}\!\times\!n_{t}}$, $b_t\in \mathbb{R}^{n_{t}\!\times\!1}$ are the learnable parameters in $c_{t}$ with kernel size $h_t\!\times\!w_t$, $n_{t-1}$ and $n_{t}$ are the sizes of input and output channels of $c_{t}$, respectively. 

\subsection{Augmentation Pathways (AP)}
\label{ssec:basic}

\begin{figure}[t]
  \centering
  %\vskip -0.4cm
  \includegraphics[width=0.6\columnwidth,page=2]{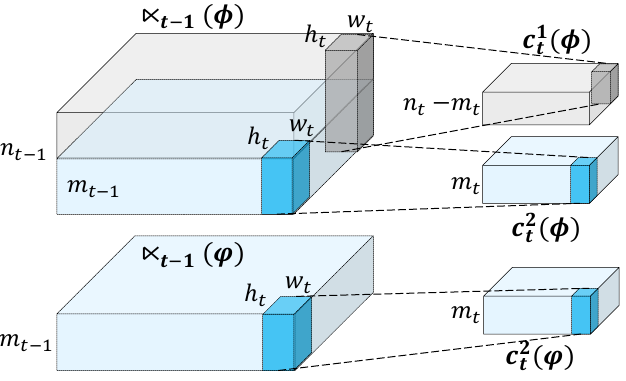}
  %\vskip -0.2cm
  \caption{The detailed structure of basic augmentation pathway based convolutional layer.}
  \label{fig:basicnestedconv}
  %\vskip -0.5cm
%\end{figure}
\end{figure}

We first introduce convolutional operations with augmentation pathways (AP-Conv), the basic unit of our proposed AP network architecture. Different from the standard convolution $c_{t}$ ($t=1,...,T$, denoting the layer index), AP version convolution $\mathbb{n}_{t}$ consists of two convolutions $c^1_{t}$ and $c^2_{t}$. $c^1_{t}$ is equipped in the main pathway, learning feature representations of lightly augmented input $\phi$ (with similar distributions with original images). $c^2_{t}$ is the pathway to learn shared visual patterns between lightly augmented image $\phi$ and heavily augmented image $\varphi$. $\varphi$ varies from different data augmentation policies, and differs from the original original image distribution. The operations of a basic AP-Conv $\mathbb{n}_{t}$ can be defined as:
\newcommand{\concat}{\ensuremath{+\!\!\!\!+\,}}
%\begin{equation}
%\setlength\abovedisplayskip{3pt}%shrink space
%\setlength\belowdisplayskip{3pt}
\begin{align}
    \mathbb{n}_{t}(\phi_i)&=c^1_{t}(\phi_i)\concat c^2_{t}(\phi_i)\nonumber\\
    &= \left(W^1_t \mathbb{n}_{t-1}(\phi_i)\!+\!b^1_t\right)\concat \left(W^2_t c^2_{t-1}(\phi_i)\!+\!b^2_t\right),\nonumber\\
    \mathbb{n}_{t}(\varphi_i) &= c^2_{t}(\varphi_i) = W^2_t c^2_{t-1}(\varphi_i) + b^2_t, 
\end{align}
%\end{equation}
where $\concat$ indicates the vector concatenation operation, $W^1_t\in\mathbb{R}^{n_{t-1}\!\times\!h_t\!\times\!w_t\!\times\!(n_t\!-\!m_t)}$, $b^1_t\in \mathbb{R}^{(n_t-m_t)\!\times\!1}$ and $W^2_t\in\mathcal{R}^{m_{t-1}\!\times\!h_t\!\times\!w_t\!\times\!m_t}$, $b^2_t\in \mathbb{R}^{m_t\!\times\!1}$ represent the convolutional weights and biases of $c^1_{t}$ and $c^2_{t}$ respectively. $m_{t-1}$ and $m_t$ denote the numbers of input and output channels of $\mathbb{n}_{t}$ for processing heavily augmented inputs and lightly augmented inputs jointly, which is smaller than $n_t$. For light augmentation inputs, the output size of $\mathbb{n}_{t}$ is same with $c_t$. As shown in Fig.~\ref{fig:basicnestedconv}, AP-Conv contains two different neural pathways inner one neural layer for $\phi$ and $\varphi$ respectively.

\newcommand{\tabincell}[2]{\begin{tabular}{@{}#1@{}}#2\end{tabular}}
\begin{table*}[t]
    \small
    \centering
    \renewcommand{\arraystretch}{1.3}
    \caption{Examples of data augmentations with their hyperparameters. Gray, Blur, Gridshuffle, MPN are manually designed heavy augmentations. RandAugment is a searched augmentation combination including 14 different image transformations (e.g., Shear, Equalize, Solarize, Posterize, Rotate. Most of them are heavy transformations).}
    %\vspace{0.7em}
    \label{tab:parameters}
    %\resizebox{\textwidth}{!}{
    \begin{tabular}{|l|l|l|} 
      \hline
      Augmentation & Hyperparameter & Description \\
      %\midrule
      %\cmidrule(r){1-3}
      %\# Params (Millions) & & \\
      \hline\hline
      %Augmentation & \multicolumn{2}{c}{Top-1 Accuracy (\%)}\\
      \texttt{Gray} & \tabincell{l}{the alpha value $\alpha\in[0,1]$ of the grayscale image when overlayed\\ over the original image for $\texttt{Gray}$} & \tabincell{l}{$\alpha$ close to 1.0 means that mostly the \\new grayscale image is visible} \\\cline{1-3}
      \texttt{Blur} & \tabincell{l}{the kernel size $k$ of $\texttt{Blur}$}  & \tabincell{l}{larger $k$ leads to more blurred image} \\\cline{1-3}
      \texttt{GridShuffle} & \tabincell{c}{the number of grids $g\times g$ in image for $\texttt{GridShuffle}$}  & \tabincell{l}{larger $g$ results in smaller grid and the\\  image is destructed more drastically} \\\cline{1-3}
      %\texttt{Salt} & \tabincell{c}{the percent $p$ of all pixels replacing with salt noise}  & \tabincell{l}{larger $p$ results in more noisy image} \\\cline{1-3}
      \texttt{MPN} & \tabincell{c}{the scaling factor $s$ of pixel values for Multiplicative Noise}  & \tabincell{l}{larger $s$ results in brighter image} \\\hline\hline
      \texttt{RandAugment}~\cite{cubuk2020randaugment} & \tabincell{l}{the number $n$ of augmentation transformations to
apply sequentially, \\and magnitude $m$ for all the transformations} & \tabincell{l}{larger $n$ and $m$ results in heavier\\ augmented image}\\
      \hline
    \end{tabular}%}
%\end{wraptable}
\end{table*}

\noindent{\textbf{Comparison to Standard Convolution}} ~~A standard convolution can be transformed into a basic AP-Conv by splitting an augmentation pathway and disabling a fraction of connections. In general, the number of parameters in $\mathbb{n}_{t}$ is $\delta_t$ less than a standard convolution under same settings, where
\begin{equation}
    \delta_t = (n_{t-1}-m_{t-1})\times m_t\times h_t\times w_t.
    \label{eq:complexity}
\end{equation}
For example, if we set $m_t\!=\!\frac{1}{2}n_t$ and $m_{t-1}\!=\!\frac{1}{2}n_{t-1}$, AP-Conv only contains \textbf{75\%} parameters in the standard Conv.
% the number of parameters in AP-Conv is only \textbf{75\%} in the standard convolution.

The only additional operation in AP-Conv is a conditional statement to assign the features of $\phi$ to $c^1_{t}$ and $c^2_{t}$, or feed the features of $\varphi$ to $c^2_{t}$.

\noindent{\textbf{Augmentation Pathways based Network}} ~~The key idea of basic augmentation pathways based network is to mine the shared visual patterns between two pathways handling inputs following different distributions. A basic constraint is that the shared features should boost object classification, which is also common objective functions of two different neural pathways:
%\begin{minipage}{\displaywidth}\small
%\begin{small}
\setlength\abovedisplayskip{3pt}%shrink space
\setlength\belowdisplayskip{3pt}
\begin{align}
%\begin{split}
    \mathcal{L}_{cls} &= \sum_{i=1}^{N}\mathcal{L}\left(f_\phi(\mathbb{n}_{T}(\phi_i)), l_i\right) + \mathcal{L}\left(f_\varphi(\mathbb{n}_{T}(\varphi_i)), l_i\right) + \lambda S_i\nonumber\\
    S_i &= \sum_{t=1}^T \left\langle c^1_{t}(\phi_i), c^2_{t}(\phi_i)\right \rangle,
    \label{eq:eq4}
\end{align}
%\end{equation}
%\end{small}
%\end{minipage}
%\]
where $f_\phi$ and $f_\varphi$ are the classifiers for light and heavy augmentations respectively, %$\left\langle \right\rangle$ denote \texttt{matmul} operation, 
$S$ is a Cross Pathways Regularization item to measure the similarity of visual patterns between neural pathways. The formulation of $S$ is similar to the standard weight decay. Both of them are \textit{L2} regularization. Denoting the loss weight of standard decay as $\omega$, for all experiments in our paper, we simply set $\lambda=0.1\omega$. Minimizing $S_i$ penalizes filter redundancy in $c^1_{t}$ and $c^2_{t}$. 
%In general, highly similar filters among two pathways would reflect the visual commonality among lightly and heavily augmented views. These commonalities would result in filter redundancy between different neural pathways. After introducing $S$, redundancy can be controlled: $c_t^2$ would be responsible for learning commonality between two neural pathways, and $c_t^1$ is forced to refer to these commonalities from $c_t^2$ through the cross pathways connections.
As a result, $c^1_{t}$ focuses on learning the $\phi$-specific features. %In our internal experiments, $S_i$ is helpful in saving the network capacity and speed up convergence. 
Moreover, owing to classification losses in Eq~\ref{eq:eq4}, $c^2_t$ is expected to highlight patterns shared between $\phi$ and $\varphi$. Finally, these common visual patterns assist $f_\phi$ to classify $\phi$ correctly. During inference, we use the label with max confidence score in $f_\phi(\mathbb{n}_{T}(I_i))$ as the prediction of image $\phi\!=\!I_i$.

Notably, AP based network can be constructed by simply replacing the standard convolutional layers in typical CNN with our AP-Conv layers, as shown in Fig.~\ref{fig:intro}. In practice, the low-level features between $\phi$ and $\varphi$ can be directly shared with each other. In most cases, the performance of a typical CNN can be significantly improved by only replacing the last few standard Conv layers with AP-Conv.%, as the experimental results reported in Section~\ref{sec:performance_compare}.

\subsection{Extensions for Augmentation Pathways}
\label{ssec:extension}
%The basic augmentation pathway is designed by considering one specific setting of heavy data augmentation or searched augmentation policy. However, 
As shown in Table~\ref{tab:parameters}, some augmentation policies have several choices of hyperparameters. Deep models are usually sensitive to these hyperparameters, since different augmentation hyperparameters for the same image may lead to a wide variety of appearances. Previous methods tend to find one proper hyperparameter according to expert knowledge or automatically searching results.  %There are some methods proposed for automatically finding a proper setting for data augmentation. However, these methods usually require additional computation cost for searching hyperparameters~\cite{cubuk2019autoaugment}, and the learned augmentation policies are dataset and network architecture dependent~\cite{lim2019fast,ho2019pba}. 

We found that common visual patterns exist among augmentation policy under different hyperparameters, and the shared feature space among them usually present dependencies. For example, the shared feature learned from Blur$(k\!=\!5)$ can benefit the recognition of image with Blur$(k\!<\!5)$. For GridShuffle, some visual detail patterns learned from small grids can be reused to represent images with large grids. Thus we extend the augmentation pathways for handling augmentation policy under various hyperparameter settings. We rank the hyperparameters of augmentation according to their distribution similarities to the original training image, and then feed the images augmented with different hyperparameters into different pathways in a high-order (nested) manner. In this way, our high-order AP can gather and structure information from augmentations with various hyperparameters.%\newline%\newline %%For example, given $\texttt{GridShuffle}$ with different amounts of grids, some visual detail patterns learned from small grids can be reused in the representation learning of images with large grids. 
%The common visual patterns still exist among augmentation policy under different hyper-parameters, and the shared feature space in pairs usually present dependencies, e.g., the 
%shared features learned from GridShuffle $(g\!=\!7)$ can benefit the classification on images with GridShuffle $(g\!<\!7)$, the shared feature learned from Blur$(k\!=\!5)$ can benefit the image recognition with Blur$(k\!<\!5)$, etc.

\begin{figure}[!t]
  \centering
  \includegraphics[width=0.85\linewidth,page=1]{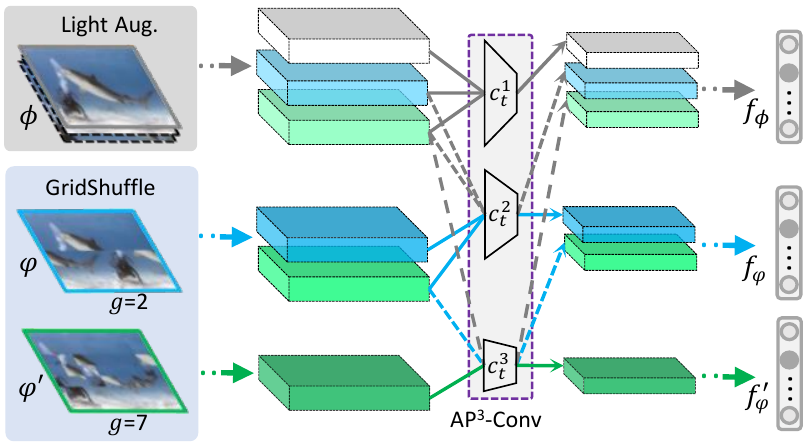}
  \caption{The 3rd-order homogeneous augmentation pathways network is extended from the basic AP but handle heavy augmentations under two different hyperparameters ($g$ for Grid Shuffle) according to the visual feature dependencies among input images.}
  \vskip -1em
\label{fig:3orderAPNet}
\end{figure}

\begin{figure*}[!t]
  \centering
  \includegraphics[width=0.92\linewidth,page=3]{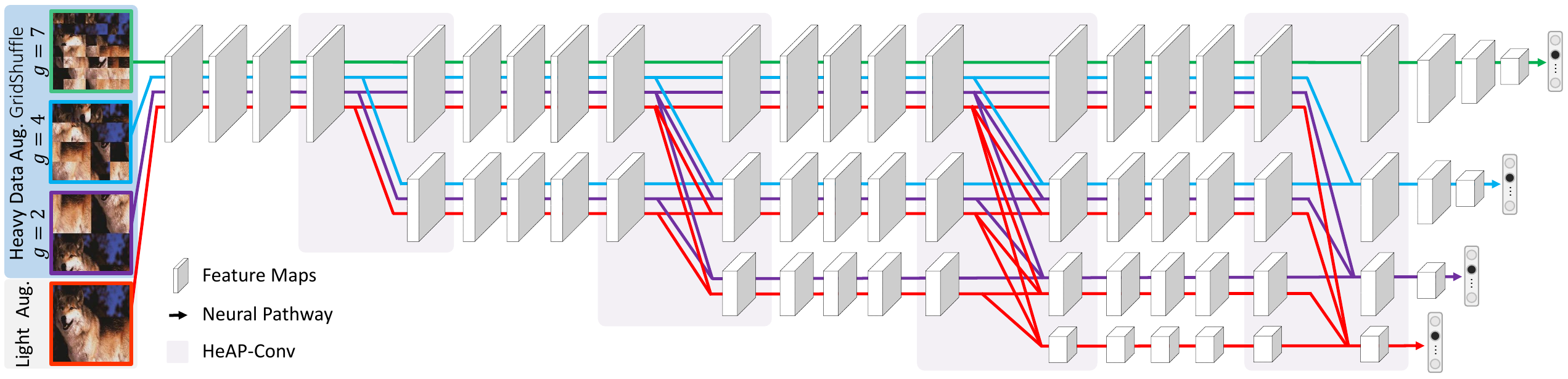}
  \caption{The network architecture of our high-order heterogeneous augmentation pathways network. Four heterogeneous neural pathways (HeAP$^4$) are responding to four different input images (lightly augmented images, GridShuffled images with g=$(2,4,7)$). Note that only the main neural pathway in red color is activated during inference.}
\label{fig:multipath}
\end{figure*}

\noindent\textbf{Extension-1: High-order Homogeneous Augmentation Pathways}\quad
We extend the basic augmentation pathway into high-order to mine shared visual patterns in different levels. Take GridShuffle as an example, we choose two different hyper-parameters to generate augmented image $\varphi\!=$ GridShuffle$(g\!=\!2)$ and $\varphi'\!=$ GridShuffle$(g\!=\!7)$. The images augmented by GridShuffle are expected to learn visual patterns inner grids, since the positions of all grids in image have been shuffled~\cite{chen2019destruction}. Considering grids in $\varphi'$ are smaller than $\phi$ and grids in $\varphi$, the local detail features learned from $\varphi'$ can be reused in $\varphi$ and $\phi$. We propose a convolution with 3rd-order homogeneous augmentation pathways (AP$^3$-Conv), which consists of three homogeneous convolutions $c^1_t$, $c^2_t$, and $c^3_t$ for handling different inputs. Similar to the basic AP-Conv, $c^1_t$ is the main augmentation pathway targeting at light augmentations $\phi$-specific feature, while augmentation pathway $c^2_t$ and $c^3_t$ are designed for learning the shared visual patterns of $\{\phi, \varphi\}$ and $\{\phi, \varphi$, $\varphi'\}$, respectively. The operation of AP$^3$-Conv can be formulated as:
\begin{equation}
\begin{split}
\mathbb{n}_{t}(\phi_i)&=c^1_{t}(\phi_i)\concat c^2_{t}(\phi_i) \concat c^3_{t}(\phi_i), \\
\mathbb{n}_{t}(\varphi_i) &= c^2_{t}(\varphi_i)\concat c^3_{t}(\varphi_i),~~~~~~~\mathbb{n}_{t}(\varphi'_i) = c^3_{t}(\varphi'_i).%= \big(W^2_t \big(c^2_{t-1}(\varphi_i)\concat c^3_{t-1}(\varphi_i)\big)  + b^2_t\big)\concat \big(W^3_t c_{t-1}(\varphi_i) + b^3_t \big), \notag\\
%\mathbb{n}_{t}(\varphi'_i) &= c^3_{t}(\varphi'_i).\notag\\% = W^3_t c^3_{t-1}(\varphi'_i) + b^3_t,\\
\end{split}
\end{equation}
%where $c_t^k(x)&= W^1_t \big(c^k_{t-1}(x)\concat ... \concat c^3_{t-1}(x)\big)  + b^k_t,\\
%c_t^1(x)&= W^1_t \big(c^1_{t-1}(x)\concat c^2_{t-1}(x)\concat c^3_{t-1}(x)\big)  + b^1_t,\\
%c_t^2(x)&=W^2_t \big(c^2_{t-1}(x)\concat c^3_{t-1}(x)\big)  + b^2_t, 1\leq k\leq 3\\\
%c_t^3(x)&=W^3_t c^3_{t-1}(x) + b^3_t\\$
In general, the standard convolution $c_t^j(x)$ can be defined as an operation filtering information from the $j$-th to the last neural pathways, 
\begin{equation}
    c_t^j(x) = W^1_t \big(c^j_{t-1}(x)\concat c^{j+1}_{t-1}(x)... \concat c^k_{t-1}(x)\big)  + b^k_t,
\end{equation}
where $1\!\leq\!j\!\leq\!k$, $k$ is the count of neural pathways in total. For AP$^3$-Conv, we set $k\!=\!3$. $c^1_{t}$ takes the outputs of $c^1_{t-1}$, $c^2_{t-1}$, $c^3_{t-1}$ as inputs, while $c^2_{t}$ takes the outputs of $c^2_{t-1}$, $c^3_{t-1}$ as inputs. In this way, the dependency across $\phi$, $\varphi$ and $\varphi'$ can be built. %Similar to the basic augmentation pathway network, the network high-order is constructed by replacing the traditional convolutional layer with our proposed AP$^K$-Conv.
Fig.~\ref{fig:3orderAPNet} indicates a network with 3rd-order homogeneous augmentation pathways (AP$^{3}$) handling two different hyperparameters for GridShuffle, whose objective function is defined as:
%where $W^1_t\in\mathbb{R}^{n_{t-1}\times h_t\times w_t\times (n_t-m_t-m'_t)}$, $b^1_t\in \mathbb{R}^{(n_t-m_t-m'_t)\times 1}$, $W^2_t\in\mathcal{R}^{(m_{t-1}+m'_{t-1})\times h_t\times w_t\times m_t}$, $b^2_t\in \mathbb{R}^{m_t\times 1}$, and $W^3_t\in\mathcal{R}^{m'_{t-1}\times h_t\times w_t\times m'_t}$, $b^3_t\in \mathbb{R}^{m'_t\times 1}$ indicate the convolutional weight and bias of $c^1_{t}$, $c^2_{t}$ and $c^3_{t}$ respectively. $m'_{t-1}$ and $m'_t$ denote the number of input and output channels of $\mathbb{n}_{t}$ by given $\varphi'$. Note that $c^1_{t}$ takes the outputs of $c^1_{t-1}$, $c^2_{t-1}$, $c^3_{t-1}$ as inputs, while $c^2_{t}$ takes the outputs of $c^2_{t-1}$, $c^3_{t-1}$ as inputs. %In this way, the dependency across $\mathbb{n}_t{\phi}$, $\mathbb{n}_t{\varphi}$ and $\mathbb{n}_t{\varphi'}$ can be built. 
%The pipeline of 3-order nested convolution is shown in Figure~\ref{fig:highorder-nc}. 
%Similar to the basic nested convolutional network, 
%The objective function of the 3rd-order augmentation pathways based network is defined as:
%\begin{small}
\setlength\abovedisplayskip{3pt}%shrink space
\setlength\belowdisplayskip{3pt}
\begin{align}\label{eq:eq7}
 %\begin{split}
    \mathcal{L}_{cls} = \sum_{i=1}^{N}&\mathcal{L}\left(f_\phi(\mathbb{n}_{T}(\phi_i)), l_i\right) + \mathcal{L}\left(f_\varphi(\mathbb{n}_{T}(\varphi_i)), l_i\right) \nonumber\\
    &+ \mathcal{L}\left(f_\varphi'(\mathbb{n}_{T}(\varphi'_i)), l_i\right) +\lambda S_i, \\
    S_i = \sum_{t=1}^T &\left\langle c^1_{t}(\phi_i), c^2_{t}(\phi_i), c^3_{t}(\phi_i)\right \rangle + \left \langle  c^2_{t}(\varphi_i), c^3_{t}(\varphi_i)\right \rangle.\nonumber
 %\end{split}
\end{align}
%\end{small}
%where $\text{sim}()$ measures the similarity among the outputs from all three sub-convolution layers for $\phi$, and the similarity among the outputs of $c^2_t$ and $c^3_t$ for $\varphi$.

The original image $\phi\!=\!I_i$ is predicted by $f_\phi(\mathbb{n}_{T}(I_i))$ during inference. 

By analogy, we can design higher-order augmentation pathways network of $k$ different homogeneous dataflow pathways, for handling $k\!-\!1$ different settings of a given heavy data augmentation policy. In general, our high-order AP$^k$-Conv can handle various settings of the given augmentation and collect useful visual patterns in different levels. At last, all features are integrated in a dependency manner and results in well-structured feature space for original image classification.%\newline

%$F(\texttt{GridShuffle}(g=7))\subset F(\texttt{GridShuffle}(g=2)) \subset F(\texttt{GridShuffle}(g=1))$, $F(\texttt{Blur}(k=5))\subset F(\texttt{Blur}(k=2)) \subset F(\texttt{Blur}(k=0))$, $F(\texttt{Gray}(\alpha=1.0))\subset F(\texttt{Gray}(\alpha=0.5)) \subset F(\texttt{Gray}(\alpha=0))$, where $F$ denotes the feature space learned from given inputs.

% For simplicity, we extend two sub-convolutions in the basic  present our approach in the context of multiclass image recognition, even
%though it can be easily generalized to other scenarios, such
%as dense image segmentation.

\noindent\textbf{Extension-2: High-order Heterogeneous Augmentation Pathways}\quad
We have adapted homogeneous neural pathways and loss functions for various hyperparameters of given heavy data augmentation in a high-order augmentation pathway network. The basic structure and settings (e.g., kernel sizes, strides in each sub-convolutional layer) of these neural pathways are the same in AP$^{k}$. However, images augmented using different hyperparameters may have different characteristics, which is a reasonable motivation for customizing the basic settings of neural pathways for inputs with different properties. Again we take GridShuffle as an example, higher-resolution representations are more suitable for learning from detailed features in smaller grids. It means that the neural pathway consists of convolutions with larger feature map outputs that would be more friendly to GridShuffle with a larger $g$.

Here we introduce another high-order extension of basic augmentation pathways for integrating representations learned from heterogeneous augmentation pathways for different characteristics. Fig.~\ref{fig:multipath} shows the pipeline of a 4th-order heterogeneous augmentation pathways (HeAP$^4$) based network with heavy augmentation in three different settings GridShuffle$(g\!=\!2,4,7)$. Similar to the architecture of HRNet~\cite{SunXLW19,WangSCJDZLMTWLX19}, different neural pathways are configured with convolutions with different kernel sizes and channel sizes and result in feature maps in different resolutions. The augmentation pathway in green color is shared among all pathways since detailed visual patterns inner grids of GridShuffle$(g\!=\!7)$ is useful for the classification of all other inputs. Four-resolution feature maps are fed into the main pathway in a nested way during inference of the original image. We apply convolution-based \textit{downsample} for zooming out the feature maps to its dependent ones. Our heterogeneous neural pathway based convolutions are used for integrating features learned from different augmentations. Each neural pathway is followed by one specific classification head. The objective function of HeAP$^4$ network is the same as the 4th-order homogeneous augmentation pathways network.

%the properties of augmented images by using different hyperparameters may be  

% Model ensemble, HR-Net
%Now, take it a step further and rethinking the 

%Let's rethinking the challenge of handling different hyperparameter settings for given destructive augmentations, 

%a naive solution is to train classification neural network for each setting, and then ensemble the outputs of all these models. However, thus such an approach is impractical since the computation cost of inferences on multiple models and results ensemble accumulates exponentially as the single model testing.

%% file: pami_src/experiments.tex
\begin{table*}[!ht]
    \renewcommand{\arraystretch}{1.3}
    \centering
    \caption{The performance on ImageNet / \#Parameters / MACs on ResNet, iResNet, ResNeXt, MobileNet V2, ConvNeXt and their basic Augmentation Pathways (AP) version on given additional heavy augmentation policy RandAugment (for generating $\varphi$). \textit{repro}: our reproduction of  each method with their original augmentation settings.}
    \label{tab:randaug}
    \begin{tabular}{|l|l|c|c|c|c|c|}
    \hline
    Metrics & Method & ResNet-50 & ResNeXt-50 32x4d & MobileNetV2 & iResNet-50 & ConvNeXt-Tiny \\\hline\hline
    & \textit{repro.} & 25.6M & 25.0M & 3.5M & 25.6M & 28.6M\\\cline{2-7}
    \multirow{-2}{*}{\#Params.} & w/ AP & \textbf{21.8M} & \textbf{21.4M} & \textbf{3.3M} & \textbf{21.8M} & \textbf{25.5M}\\\hline\hline
    & \textit{repro.} & 4.11G & 4.27G & 0.32G & 4.15G & 4.47G \\\cline{2-7}
    \multirow{-2}{*}{MACs} & w/ AP & \textbf{3.91G} & \textbf{4.06G} & \textbf{0.30G} & \textbf{3.95G} & \textbf{4.30G}\\\hline\hline
    & \textit{repro.} & 76.19 / 93.13 & 77.48 / 93.66 & 71.97 / 90.37 & 77.59 / 93.55 & 81.98 / 95.88\\
    & w/ $\varphi$ & 77.12 / 93.45 & 77.67 / 93.76 & 72.04 / 90.38 & 77.20 / 93.52 & 81.56 / 95.75 \\\cline{2-7}
    \multirow{-3}{*}{Acc.(\%)} &w/ AP & \textbf{77.97} / \textbf{93.92} & \textbf{78.18} / \textbf{94.07} & \textbf{72.34} / \textbf{90.48} & \textbf{78.20} / \textbf{93.95} & \textbf{82.23} / \textbf{96.01} \\\hline
    \end{tabular}
\end{table*}
\section{ImageNet Experiments and Results}
We evaluate our proposed method on ImageNet~\cite{deng2009imagenet} dataset (ILSVRC-2012), due to its widespread usage in supervised image recognition. %There are 1.28 million images in 1000 classes for training and $50,000$ images for validation. 
Since the main purpose of data augmentation is to prevent overfitting, we also construct two smaller datasets from the training set of ImageNet by randomly sampling 100 and 20 images for each class, named ImageNet$_{100}$ and ImageNet$_{20}$. ImageNet$_{100}$ is also used for ablation studies in this paper. 

We apply augmentation pathways on \textbf{six} widely used backbone networks covering typical ConvNet developments from 2015 to 2022, including:
\begin{itemize}[itemsep=2pt,topsep=0pt,parsep=0pt]
\item \textbf{ResNet}~\cite{he2016deep} (2015), stacking residual and non-linear blocks.
\item \textbf{ResNeXt}~\cite{xie2017aggregated} (2017), repeating blocks that aggregates a set of transformations with the same topology.
\item \textbf{MobileNetV2}~\cite{sandler2018mobilenetv2} (2018), mobile architecture based on the inverted residuals and linear bottlenecks.
\item \textbf{HRNet}~\cite{WangSCJDZLMTWLX19} (2019), exchanging information across steams with different resolutions.
\item \textbf{iResNet}~\cite{duta2020improved} (2020), using ResGroup blocks with group convolutional layers, improved information flow and projection shortcut.
\item \textbf{ConvNeXt}~\cite{liu2022convnet} (2022), designed for ``modernizing'' ConvNet toward the design of a vision Transformer (\textit{e.g.} Swin-T).
\end{itemize}
%single branch styled ResNet-50, ResNet-101, its improved version iResNet-50, group convolution network ResNeXt, inverted residual structure MobileNetV2, multi-scale
%heterogeneous network HRNet and the most advanced convolutional network ConNeXt with Transformer-style design, covering the period of CNN development from 2015 to 2022. %All experimental settings (e.g., initial learning rate, optimization method, learning rate decay approach) follow the original paper  literatures of~\cite{he2016deep,duta2020improved}. 
Single central-crop testing accuracies on the ImageNet validation set are applied as the evaluation metric for all experiments.% results in this paper.

\subsection{Implementation Details}
Following standard practices~\cite{he2016deep,krizhevsky2012imagenet,hu2018squeeze}, we perform standard (light) data augmentation with random cropping 224$\times$224 pixels and random horizontal flipping for all baseline methods except ConvNeXt. Same with the original setting of ConvNeXt~\cite{liu2022convnet} training implementation\footnote{\url{https://github.com/facebookresearch/ConvNeXt}}, we adopt schemes including \textit{Mixup}, \textit{Cutmix}, \textit{RandAugment}, and \textit{Random Erasing} as the light augmentations policies for ConvNeXt models. All other hyperparameters are consistent with each method's default settings.
%Following standard practices~\cite{he2016deep,krizhevsky2012imagenet,hu2018squeeze}, we perform standard (light) data augmentation with random cropping 224$\times$224 pixels and random horizontal flipping for baseline methods. %For all classification models in our experiments, the optimization is performed using SGD with momentum 0.9 and a minibatch size of 256. The initial learning rate is set to 0.1 and decreased by a factor of 10 every 30 epochs. 
The augmentation pathways version of baseline methods is designed by replacing all standard convolutional layers in the last stage~\cite{he2016deep,hu2018squeeze} (whose input size is $14\!\times\!14$, and output feature map size is $7\!\times\!7$) by AP$^k$-Conv. We set the input and output channel sizes of each sub-convolution $c^1, c^2, ..., c^k$ in AP$^k$-Conv as $1/k$ of the input and output channel size in the replaced standard convolutional layer, respectively. For architectures containing group convolution layers, \textit{e.g.} ResNeXt, MobileNetV2 and ConvNeXt, we remain the number of groups of each convolution inner every AP$^k$-Conv to be the same with its corresponding original group convolution layer. For HeAP networks, we equip heterogeneous augmentation pathways after each stage. %For all experiments, $\lambda$ is set as $1e^{-7}$.
More implementation details can be found in our released source code\footnote{\url{https://github.com/ap-conv/ap-net}}.

\subsection{Performance Comparison}\label{sec:performance_compare}
%\textbf{AP on Searched Augmentations}\quad 
%Following the standard settings of RandAugment and its PyTorch implementation\footnote{https://github.com/ildoonet/pytorch-randaugment},
Following the settings of other heavy augmentation related works~\cite{wang2022contrastive,bai2022directional}, we firstly apply RandAugment with hyperparameter $m\!=\!9$, $n\!=\!2$ for generating heavy augmented view $\varphi$. The experimental results on different network architectures are reported in Table~\ref{tab:randaug}. Our proposed
AP consistently benefits all these ConvNets with fewer model parameters and lower inference computational cost. It can be found that the RandAugment policy searched for ResNet-50 architecture results in a performance drop on iResNet-50\footnote{https://github.com/iduta/iresnet}. While our augmentation pathways stably improve all architectures. The performance improvement of MobileNetV2 w/ AP is not as significant as the results on other architectures. It is mainly due to the limited parameters of MobileNetV2 bounded its feature representation ability and restricted the capacity of visual patterns from various augmented views. Besides, since we apply additional RandAugment policy based on the lightly augmented view $\phi$ to generate the heavier augmented view $\varphi$ for ConvNeXt, using RandAugment twice results in performance degradation on ConvNeXt-Tiny. However, our AP can still aggregate information beneficial to the classification task from the heavier augmented view $\varphi$. These experimental results demonstrate the robustness and generality of AP.

\begin{table}[t]
    %\footnotesize
    \centering
    \renewcommand{\arraystretch}{1.3}
    \caption{Performance comparison on ImageNet subsets. AP-ResNet achieves significant improvements with different heavy data augmentation policies.}
    \label{tab:tiny_dataset}
    %\begin{center}
    %\setlength{\tabcolsep}{1.7mm}{
    \begin{tabular}{|l|l|c|c|} 
      \hline
      %s\multicolumn{6}{l}{ImageNet-2012 labels: 1000} \\
      %\multicolumn{2}{|l}{\#Image for each class:} &  \multirow{2}{*}{100 ($\texttt{ImageNet}_{100}$)} & \multirow{2}{*}{20 ($\texttt{ImageNet}_{20}$)} \\
      %\cline{1-6}
      Augmentation & Model & ImageNet$_{100}$ & ImageNet$_{20}$\\
      %\multicolumn{2}{c}{Part}                   \\
      %\cmidrule(r){1-2}
      \hline\hline
      Random Crop,Flip & ResNet & 45.01 / 70.04 & 9.59 / 23.75 \\
            \hline\hline
      \multirow{2}{*}{GridShuffle} & ResNet & 43.95 / 68.97 & 9.88 / 23.81 \\
      & AP-ResNet & \textbf{45.62}      / \textbf{70.93} & \textbf{11.53} / \textbf{27.85} \\\cline{1-4}
      \multirow{2}{*}{MPN} & ResNet & 45.51 / 70.78 & 10.64 / 25.36 \\
      & AP-ResNet & \textbf{46.98} / \textbf{71.64} & \textbf{11.14} / \textbf{26.57} \\\cline{1-4}
      \multirow{2}{*}{Gray} & ResNet & 45.83 / 71.08 & 9.63 / 24.49 \\
      & AP-ResNet & \textbf{46.83} / \textbf{72.01} & \textbf{11.68} / \textbf{27.85} \\\hline\hline
      %\multirow{2}{*}{AutoAug} & ResNet & 53.47 / 76.50 & 22.75 / 44.26 \\
      %& AP-ResNet & \textbf{40.73} / \textbf{64.60} & \textbf{11.55} / \textbf{26.52} \\\cline{1-4}
      \multirow{2}{*}{RandAugment} & ResNet & 51.75 / 75.66 & 17.59 / 37.06  \\
      & AP-ResNet & \textbf{53.74} / \textbf{76.83} & \textbf{20.80} / \textbf{40.86} \\%\cline{1-4}
      \hline
    \end{tabular}%}
    %\end{center}
\end{table}

\iffalse
\begin{table*}[t]
\renewcommand{\arraystretch}{1.3}
    \centering
    \caption{The recognition accuracy of the proposed 3rd-order augmentation pathway (AP$^3$) based ResNet-50 by equipping additional augmentation GridShuffle with different hyperparameters.}
    \label{tab:highorder_res}
    %\begin{center}
    %\setlength{\tabcolsep}{1.3mm}{
    %\setlength{\tabcolsep}{0.5mm}{
    \begin{tabular}{|l|c|c|l|c|c|} 
      \hline
      %s\multicolumn{6}{l}{ImageNet-2012 labels: 1000} \\
      Method & \#Para & MACs & Augmentation & ImageNet$_{100}$ & ImageNet \\
      %\cmidrule(r){1-5}
      %Additonal Argumentation & Model & Top-1 & Top-5 & Top-1 & Top-5\\
      %\multicolumn{2}{c}{Part}                   \\
      %\cmidrule(r){1-2}
      \hline\hline
      ResNet & 25.6M & 4.11G & Baseline & 45.01 / 70.04 & 75.49 / 92.63 \\\cline{1-6}
      \multirow{2}{*}{AP-ResNet} & \multirow{2}{*}{21.8M} & \multirow{2}{*}{3.91G} & GridShuf($g$=2) & 47.91 / 72.80 & 76.21 / 93.04 \\
      & & & GridShuf($g$=7) & 45.62 / 70.93 & 75.76 / 92.71 \\\cline{1-6}
      AP$^3$-ResNet & \textbf{20.6M} & \textbf{3.84G} & GridShuf($g$=2,7) & \textbf{48.11 / 72.91} & \textbf{76.52 / 93.24} \\
      \hline
    \end{tabular}%}    
    %\end{center}
\end{table*}
\fi

\noindent\textbf{AP on Fewer Labels}\quad 
We also applied augmentation pathways in small datasets ImageNet$_{100}$ and ImageNet$_{20}$ to test on the practical scenario of data scarcity. We selected three manually designed heavy data augmentations GridShuffle$(g\!=\!7)$, Gray$(\alpha\!=\!1)$, MPN$(s\!=\!1.5)$ and RandAugment$(m\!=9,n\!=2)$ besides light augmentations. The experimental results are reported in Table~\ref{tab:tiny_dataset}. We can find that AP-Net significantly boosts the performance on small datasets. Note this is practically useful when training data is expensive to obtain.

\noindent\textbf{High-order Homogeneous Augmentation Pathways}\quad 
In Table~\ref{tab:highorder_res}, we compare the results from the standard ResNet-50, its basic AP version, and 3rd-order version AP$^3$. %We apply GridShuffle in our experiments since it is one of the most representative heavy data augmentations. 
In detail, our 3rd-order augmentation pathway is designed for adapting two RandAugment with different hyper-parameters. We find that AP$^3$ can further improve the performance of the 2nd-order basic AP-Conv based network. The significant gains as introducing more different hyper-parameters indicate that structuring the subdivision of generalities among different features spaces in a dependent manner benefits the object recognition.

\begin{figure*}[t]
\begin{minipage}[t!]{.62\linewidth}
%\begin{figure}[t]
\centering
\renewcommand{\arraystretch}{1.3}
    \footnotesize
    \captionof{table}{Recognition accuracy of: 1) 3rd-order augmentation pathway (AP$^3$) based ResNet-50 by equipping additional augmentation RandAugment$^2$$((n,m) \in\{(1,5), (2,9)\})$, and 2) heterogeneous augmentation pathways (HeAP$^4$) based network by equipping additional augmentation RandAugment$^3$$((n,m) \in\{(1,5), (2,9), (4,15)\})$.}
    \label{tab:highorder_res}
    %\begin{center}
    %\setlength{\tabcolsep}{1.3mm}{
    %\setlength{\tabcolsep}{0.5mm}{
    \begin{tabular}{|l|c|c|l|c|c|} 
      \hline
      %s\multicolumn{6}{l}{ImageNet-2012 labels: 1000} \\
      Method & \#Params. & MACs & Augmentation & ImageNet$_{100}$ & ImageNet \\
      %\cmidrule(r){1-5}
      %Additonal Argumentation & Model & Top-1 & Top-5 & Top-1 & Top-5\\
      %\multicolumn{2}{c}{Part}                   \\
      %\cmidrule(r){1-2}
      \hline\hline
      \multirow{2}{*}{ResNet~\cite{he2016deep,cubuk2020randaugment}} & \multirow{2}{*}{25.6M} & \multirow{2}{*}{4.11G} & Baseline & 45.01 / 70.04 & 76.64 / 93.24\\\cline{4-6}
      & & & RandAugment$^2$ & 51.67 / 75.45  & 77.03 / 93.41\\\hline
      AP-ResNet & 21.8M & 3.91G & RandAugment$^2$ & 53.58 / 76.61 & 77.59 / 93.68\\\hline
      AP$^3$-ResNet & \textbf{20.6M} & \textbf{3.84G} & RandAugment$^2$ &\textbf{54.08} / \textbf{77.11} & \textbf{78.06} / \textbf{93.92} \\
      \hline\hline
      \multirow{2}{*}{HRNet~\cite{WangSCJDZLMTWLX19}} & \multirow{2}{*}{67.1M} & \multirow{2}{*}{14.93G} & Baseline & 51.53 / 75.58 & 78.81 / 94.41 \\\cline{4-6}
      & & & RandAugment$^3$ & 53.52 / 77.54 & 77.28 / 93.95\\\hline
      HeAP$^4$-HRNet & \textbf{59.9M} & \textbf{13.97G} & RandAugment$^3$ & \textbf{54.35} / \textbf{78.24} & \textbf{79.25} / \textbf{94.78} \\
      \hline
    \end{tabular}
\end{minipage}
~~~
\begin{minipage}[t!]{.37\linewidth}
\centering
        \renewcommand{\arraystretch}{1.3}
        \centering
        \footnotesize
        %\vspace{-5pt}
        \captionof{table}{AP-ResNet-50 w/o sharing weights for GridShuffle(7).}
        %\vskip -1.8em
        %\vskip -0.7em
        \label{tab:irrelevant}
        \begin{tabular}{|l|c|c|c|}
        \hline
        $m_t=$ & $\frac{1}{2}n_t$ & $\frac{2}{3}n_t$ & $n_t$ \\\hline\hline
        %Acc.(\%) & 45.62 / 70.93 & 45.55 / 70.78 & 43.95 / 68.97 \\\hline
        Acc. &  45.59 \scriptsize$\pm$0.13 & 45.53 \scriptsize$\pm$0.11 & 43.95 \scriptsize$\pm$0.11 \\\hline
        \end{tabular}
        \vspace{1.5em}
        ~\\
        %\vskip 2em
        %\vspace{2em}%\vskip 2em
        \includegraphics[width=0.82\linewidth, page=1]{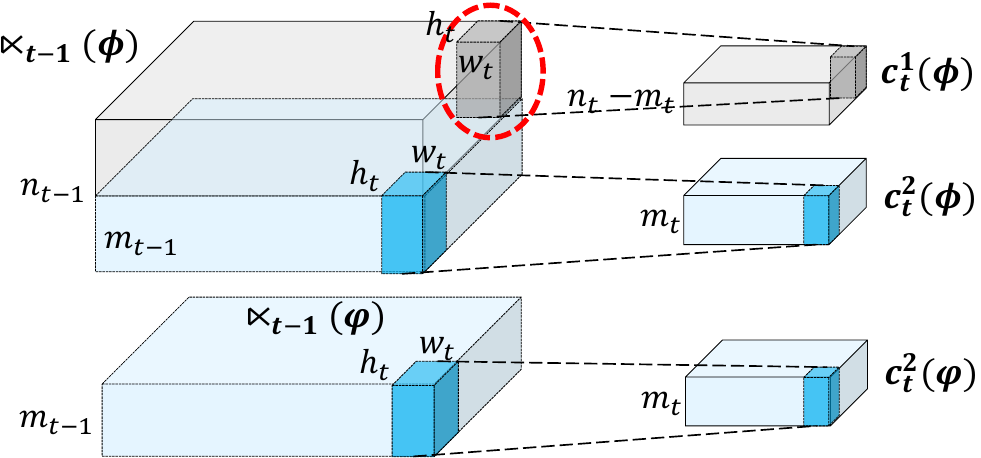}\label{fig:wosharing}
        \vspace{0.8em}
        \captionof{figure}{The structure of augmentation pathway based convolutional layer without sharing feature.}\vspace{-2em}
\end{minipage}
\end{figure*}

\noindent\textbf{High-order Heterogeneous Augmentation Pathways}\quad 
Following the framework described in Fig.~\ref{fig:multipath}, we adapt an HRNet-W44-C~\cite{WangSCJDZLMTWLX19} style network architecture for 4th-order heterogeneous augmentation pathways network by replacing all multi-resolution convolution with HeAP$^4$-Conv. Unlike the HRNet, which can only pass one image once, its HeAP$^4$ variant can handle four different inputs simultaneously. The hierarchical classification head of HRNet is disabled in HeAP$^4$. Four parallel loss functions follow four different neural pathways in HeAP$^4$-HRNet. Only the neural pathway for lightly augmented inputs is activated during inference. Table~\ref{tab:highorder_res} summarizes the classification results of HRNet and our HeAP$^4$-HRNet. HeAP$^4$-HRNet significantly outperforms HRNet on ImageNet$_{100}$ with fewer parameters and lower computational cost. Recall that HeAP$^4$-HRNet and HRNet are two different architectures due to the completely different data flow, HeAP convolutional layers, and classification heads.

\subsection{Discussions}
To evaluate the statistical significance and stability of the proposed method, we report the mean and standard deviation of the accuracy from five trials for all below ablation experiments on ImageNet$_{100}$. %In special, we compute a model's error rate as the median error of the final five epochs and report the mean and standard deviation (std) of the error from five independent runs.

\noindent\textbf{Impact of the Cross Pathways Connections}\quad
We design ablation studies by removing cross-pathways connections (w/o feature sharing among pathways) in AP-Conv but remaining the loss functions in Eq.~(\ref{eq:eq4}) and Eq.~(\ref{eq:eq7}) (as shown in Fig.~\ref{fig:wosharing}). For standard ConvNet, heavily augmented views can directly influence the training of all parameters. However for AP-Net w/o sharing weights, heavily augmented views can only affect a half set of parameters' training (if we set $m_t\!=\!\frac{1}{2}n_t$ as default).

\iffalse
\begin{figure}[h]
\centering
%\vspace{0pt}
\includegraphics[width=0.7\linewidth, page=1]{rebuttal/withoutsharefea_v1.pdf}
%\vskip -0.7em
\caption{The structure of augmentation pathway based convolutional layer but without sharing feature.}
%\vskip -1.8em
\label{fig:wosharing}
%\end{figure}
\end{figure}%

\begin{table}[t]
\renewcommand{\arraystretch}{1.3}
\centering
%\vspace{-5pt}
\caption{AP-ResNet-50 w/o sharing weights for GridShuffle(7).}
    %\vskip -1.8em
    \label{tab:irrelevant}
    %\setlength{\tabcolsep}{0.4mm}{
    \begin{tabular}{|l|c|} 
      \hline
      %s\multicolumn{6}{l}{ImageNet-2012 labels: 1000} \\
      $m_t=$ & ImageNet$_{100}$\\\hline\hline
      $\frac{1}{2}n_t$ & 45.62 / 70.93 \\
      $\frac{2}{3}n_t$ & 45.55 / 70.78 \\
      $n_t$ & 43.95 / 68.97 \\\hline
    \end{tabular}
    %\vskip 1em
\end{table}
\fi

The results in Table~\ref{tab:lightlyconn} show that (1) our proposed \textbf{loss function} leads to +0.87\% improvement over baselines, and (2) \textbf{AP-style architecture} further boost 1.18\% gain, due to the visual commonality learned among pathways.

Moreover, Table~\ref{tab:irrelevant} shows that increasing the influence of heavily augmented views leads to performance drop (ConvNet is equal to AP-Net w/o sharing weight when $m_t=n_t$). Such phenomenon is owing to the irrelevant feature bias introduced by the heavy augmentations. The divided pathways design can suppress such irrelevance.% and results in higher accuracy.

\begin{table}[t]
\renewcommand{\arraystretch}{1.3}
    \centering
    \caption{The effect of removing cross pathways connections, and randomly feeding inputs to different pathways. Heavy augmentation is RandAugment.}
    \label{tab:lightlyconn}
    %\begin{center}
    \setlength{\tabcolsep}{1.5mm}{
    \begin{tabular}{|l|c|} 
      \hline
      %s\multicolumn{6}{l}{ImageNet-2012 labels: 1000} \\
      Method & ImageNet$_{100}$ \\
      %\cmidrule(r){1-5}
      %Additonal Argumentation & Model & Top-1 & Top-5 & Top-1 & Top-5\\
      %\multicolumn{2}{c}{Part}                   \\
      %\cmidrule(r){1-2}
      \hline\hline
      ResNet-50 & 51.69 \scriptsize{$\pm$0.09} \\\hline
      AP-ResNet-50 w/o sharing feature & 52.58 \scriptsize$\pm$0.11 \\\hline
      AP-ResNet-50 w/ randomly input & 52.80 \scriptsize$\pm$0.14 \\\hline
      AP-ResNet-50 & 53.76 \scriptsize$\pm$0.08  \\\hline
      %\multirow{2}{*}{GridShuffle(2,7)} &AP$^3$-ResNet-50 w/o sharing feature & 46.54 / 71.45 \\\cline{2-3}
      %& AP$^3$-ResNet-50 &  48.11 / 72.91    \\\hline
    \end{tabular}}    
    %\end{center}
\end{table}

\iffalse
\noindent\textbf{Impact of Augmentation Grading:}\quad
We also report the results using high-order AP with same augmentation for different pathways in Table~\ref{tab:sameAug}. The results show that considering the visual dependency among different augmentation settings though AP-Conv leads to .

\begin{table}[t]
    \small
    \begin{center}
    %\setlength{\tabcolsep}{0.4mm}{
    \begin{tabular}{|l|l|c|} 
      \hline
      %s\multicolumn{6}{l}{ImageNet-2012 labels: 1000} \\
      Method & Augmentation & ImageNet$_{100}$ \\
      %\cmidrule(r){1-5}
      %Additonal Argumentation & Model & Top-1 & Top-5 & Top-1 & Top-5\\
      %\multicolumn{2}{c}{Part}                   \\
      %\cmidrule(r){1-2}
      \hline\hline
      \multirow{2}{*}{ResNet-50} & GridShuffle(g=7) & 43.95 / 68.97 \\\cline{2-3}
      & GridShuffle(g=2) &  /  \\\hline%\hline
       & GridShuffle(g=2,7) &  \\\hline%\hline
      %AP-ResNet-50 & GridShuffle(g=7) & 45.62 / 70.93 \\\hline
      \multirow{3}{*}{AP$^3$-ResNet-50 } & GridShuffle(g=7,7) & 45.55 / 70.78 \\\cline{2-3}
      & GridShuffle(g=2,2) & 46.97 / 71.62 \\\cline{2-3}
      & GridShuffle(g=2,7) & 48.11 / 72.91 \\\hline
    \end{tabular}%}    
    \end{center}
    \caption{Ablation study of using same augmentation.}
    \label{tab:sameAug}
\end{table}
\fi

\begin{figure}
    \centering
\includegraphics[width=0.75\linewidth, page=1]{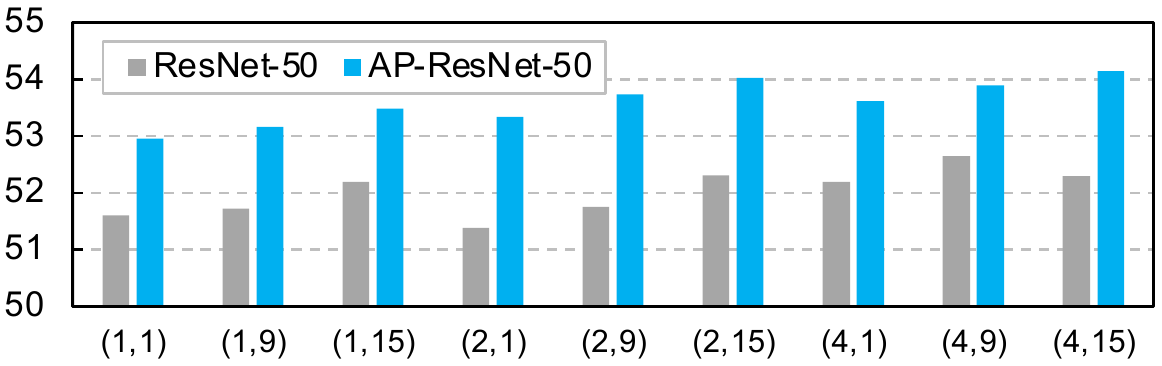}
%\vskip -0.7em
\caption{Top-1 accuracy (\%) on ImageNet$_{100}$ by using RandAugment with different ($n$,$m$).}
\label{fig:randaug}
\end{figure}

\noindent\textbf{Impact of Distortion Magnitudes of Augmentations}\quad
The experimental results in Fig.~\ref{fig:randaug} shows that our AP method can stably boosts the performance of ConvNet under various hyperparameters for RandAugment.

\noindent\textbf{Impact of Cross Pathways Regularization $S$}\quad To demonstrate the effects of $S$, we perform the regularization item separation experiments on AP-ResNet-50 with RandAugment. The results are shown in Table~\ref{tab:abalation}. We also compared the AP-ResNet-50 performance by applying different settings of $\lambda=n\times \omega$ for evaluating AP-Net's sensitivity to the choice of $\lambda$. It shows that cross pathways regularization benefits the feature space structure across different neural pathways, resulting in better performance. But too high loss weight for $S$ would lead to a performance drop, behaving similarly to the standard weight decay in the common neural network training.
% We can find that the cross pathways regularization can further improve the performance. %It mainly due to the constraint of $S$ can help the AP network learn more diverse filters, and save the model capacity for covering more diverse visual patterns.
\iffalse
\begin{table}[t]
\renewcommand{\arraystretch}{1.3}
    \centering
    \caption{The effect of removing cross pathways regularization term $S$ from AP-ResNet-50 with RandAugment.}
    \label{tab:abalation}
    %\begin{center}
    \begin{tabular}{|l|c|c|} 
      \hline
      %s\multicolumn{6}{l}{ImageNet-2012 labels: 1000} \\
      Method & ImageNet$_{100}$ & ImageNet$_{20}$\\\hline\hline
      %Method & Augmentation & Top-1 & Top-5 & Top-1 & Top-5 \\
      %Additonal Argumentation & Model & Top-1 & Top-5 & Top-1 & Top-5\\
      %\multicolumn{2}{c}{Part}                   \\
      %\cmidrule(r){1-2}
      AP-ResNet-50 w/o $S$ & 53.19 / 76.46 & 20.32 / 40.51 \\\hline
      AP-ResNet-50 & 53.74 / 76.83 & 20.80 / 40.86\\\hline
    \end{tabular}    
    %\end{center}
\end{table}
\fi
\begin{table}[t]
\renewcommand{\arraystretch}{1.3}
    \centering
    \caption{The impact of cross pathways regularization term $S$ and its weight for AP-ResNet-50 with RandAugment.}
    \label{tab:abalation}
    %\begin{center}
     \setlength{\tabcolsep}{1.1mm}{
    \begin{tabular}{|l|c|c|c|c|c|c|c|} 
      \hline
      $\lambda$ & $10 \omega$ & $\omega$ & $0.1 \omega$ & $0.01 \omega$ & $0$ (w/o $S$) \\\hline\hline
      Acc. & 52.86 \scriptsize$\pm$0.09 & 53.14 \scriptsize$\pm$0.08 & 53.76 \scriptsize$\pm$0.08 & 53.45 \scriptsize$\pm$0.10 & 53.19 \scriptsize$\pm$0.13 \\\hline
    \end{tabular}}
    %\end{center}
\end{table}

\noindent\textbf{Generalize the `light vs. heavy'' Augmentation Policy Settings to ``basic vs. heavier''}\quad
Inspired by the related work~\cite{bai2022directional}, defining $d$ as the deviation of augmented view from the original view, given two augmented view $\phi$ and $\varphi$, we denote $\varphi$ is heavier than $\phi$ only if $d(\varphi)>d(\phi)$. There are two situations to adjudge $d(\varphi)>d(\phi)$: 1) $\varphi$ and $\phi$ are augmented by the same policies, but $\varphi$ is augmented with more aggressive hyperparameter. 2) $\varphi$ is augmented by policies which is a \textit{proper superset} of augmentations used for generating $\phi$. In AP, the basic view $\phi$ and the heavier view $\varphi$ are fed to the main and augmentation pathway, respectively.

It means some heavy augmentation policies may generate basic view $\phi$, \textit{e.g.} ConvNeXt applies the combination of Random Crop, Mixup, Cutmix, RandAugment, and Random Erasing as basic augmentations for generating $\phi$. We can introduce another RandAugment on $\phi$ to generate heavier view $\varphi$ for ConvNeXt. The experimental results in Table~\ref{tab:randaug} show that AP-ConvNeXt-Tiny with twice RandAugment outperforms ConvNeXt-Tiny.

Accordingly, heavier view $\varphi$ can be generated by applying additional light augmentation, \textit{e.g.} we can apply another crop operation based on $\phi$ to generate the heavier view $\varphi$ (simulating the aggressive crop operation), and it still results in performance improvement, as shown in Table~\ref{tab:aggressive_crop}.
\begin{table}[t]
\renewcommand{\arraystretch}{1.3}
    \centering
    \caption{Accuracy after introducing aggressive crop operation.}
    \label{tab:aggressive_crop}
    %\begin{center}
    \setlength{\tabcolsep}{1.5mm}{
    \begin{tabular}{|l|l|c|} 
      \hline
      %s\multicolumn{6}{l}{ImageNet-2012 labels: 1000} \\
      Method & Augmentation & ImageNet$_{100}$ \\
      %\cmidrule(r){1-5}
      %Additonal Argumentation & Model & Top-1 & Top-5 & Top-1 & Top-5\\
      %\multicolumn{2}{c}{Part}                   \\
      %\cmidrule(r){1-2}
      \hline\hline
      \multirow{2}{*}{ResNet-50} & Standard Crop & 44.98 \scriptsize$\pm$0.10 \\\cline{2-3}
      & Aggressive Crop & 50.07 \scriptsize$\pm$0.12\\\hline
      AP-ResNet-50 & Aggressive Crop & \textbf{52.46} \scriptsize$\pm$0.09 \\\hline
    \end{tabular}}    
    %\end{center}
\end{table}

\noindent\textbf{Model Inference}\quad
The augmented pathways are designed to stabilize main-pathway training when heavy data augmentations are present. During inference, no heavy augmentation are adopted, only $f_\phi$ in the main neural pathway for the original image are used for computing probability.

\noindent\textbf{Model Complexity}\quad %\footnote{\#Params. and MACs reported in this paper is counted by the implementation from \url{https://github.com/Lyken17/pytorch-OpCounter}}}
Although AP usually takes more memory cost during model training than the standard ConvNet, many connections can be cut out while replacing traditional convolutions with AP-Convs. Thus the AP version of a given standard CNN network has fewer parameters (\#Params.) to learn and lower computational cost (GMACs, Multiply-Accumulate Operations) during inference, as specified in Tables~\ref{tab:randaug},~\ref{tab:highorder_res} and  Eq.~(\ref{eq:complexity})..
%The AP$^3$-ResNet-50 has fewer parameters in convolutional layers than AP-ResNet-50, but it contains more batch normalization layers than AP$^3$-ResNet-50, since augmentation pathways maintain separate batch normalization layers~\cite{ioffe2015batch}. The computation cost and parameter size of AP$^3$-ResNet-50 is roughly the same as AP-ResNet-50. 

%% file: pami_src/conclusion.tex
\section{Conclusion}\label{sec:futurework}

%We have proposed a nested sparse network, named NestedNet, to realize an n-in-1 nested structure in a neural network, where several networks with different sparsity ratios are contained in a single network and learned simultaneously. To exploit such structure, novel weight pruning and scheduling strategies have been presented. NestedNet is an efficient architecture to incorporate multiple knowledge or additional information within a neural network, while existing networks are difficult to embody such structure. NestedNets have been extensively tested on various applications and demonstrated that it performs competitively, but more efficiently, compared to existing deep architectures.

%We propose augmentation pathways for adapting neural network design to data augmentation policies. 
The core concepts of our proposed Augmentation Pathways for stabilizing training with data augmentation can be concluded as: 1) Adapting different neural pathways for inputs with different characteristics. %2) Highlighting the shared visual patterns learned from different pathways, while suppressing irrelevance irrelevant patterns from augmented images. 
2) Integrating shared features by considering visual dependencies among different inputs. %Respecting to these concepts, we also provide some other potential extensions and research directions for AP-Conv in Section~\ref{sec:futurework}. 
%Augmented images are distributed to different neural pathways by considering the visual dependency among them. %During training, the feature space is restructured by highlighting the visual commonality but suppresses irrelevance among all pathways. 
%Moreover, 
Two extensions of AP are also introduced for handling data augmentations in various hyper-parameters. In general, our AP based network is more efficient than traditional CNN with fewer parameters and lower computational cost, and results in stable performance improvement on various datasets on a wide range of data augmentation polices.%, including manually designed heavy augmentation and the searching-based augmentation. %The latent magic implied in heavy data augmentations is released. 

%In this work, we focused on applying AP-Conv to heavy data augmentation with various hyper-parameters. It is also possible to apply multiple AP-Conv in a parallel way for mining shared features between light and multiple heavy data augmentations simultaneously.

%The core concepts of the augmentation pathway can further guide neural network design for different application scenarios, e.g., 1) mining shared features across different datasets for boosting the performance each other; 2) replacing heavy data augmentations (various hyper-parameters) with the images generated from GAN~\cite{antoniou2017data} (under various settings) for robust image classification. Moreover, the AP-Conv based network design is inspired by some prior knowledge, such as the properties of heavy data augmentation, and the influences of different hyperparameters for data augmentations. To reduce dependency on prior knowledge, some meta-learning based methods maybe suitable for guiding augmentation pathways based network design.

%future directions also include: 1) meta-learning based dependent feature ensemble scheme; 2) reinforcement learning based self-adaptive neural pathway selecting for different kinds of inputs, etc.